\documentclass[10pt,twocolumn,letterpaper]{article}

\usepackage{iccv}
\usepackage{times}
\usepackage{epsfig}
\usepackage{graphicx}
\usepackage{amsmath}
\usepackage{amssymb}
\usepackage{pifont}
\usepackage{bm}
\usepackage{booktabs}
\usepackage{overpic}
\newcommand{\cmark}{\ding{51}}%
\newcommand{\xmark}{\ding{55}}%
\usepackage[absolute,overlay]{textpos}
\usepackage{multirow} 
\usepackage[dvipsnames, table]{xcolor} 
\usepackage{adjustbox} 

\DeclareMathOperator*{\argmin}{arg\,min}
\definecolor{Goldenrod}{RGB}{245,245,220}
\definecolor{mygray}{gray}{.9}
\definecolor{mypink}{rgb}{.99,.91,.95}
\definecolor{mycyan}{cmyk}{.3,0,0,0}

\usepackage{color}

\usepackage[pagebackref=false, breaklinks=true, letterpaper=true, colorlinks,
            citecolor=citecolor, linkcolor=linkcolor, bookmarks=false]{hyperref}
\definecolor{citecolor}{HTML}{0071BC}
\definecolor{linkcolor}{HTML}{ED1C24}
\usepackage[capitalize]{cleveref}
\crefname{section}{Sec.}{Secs.}
\Crefname{section}{Section}{Sections}
\Crefname{table}{Table}{Tables}
\crefname{table}{Tab.}{Tabs.}

\iccvfinalcopy 
\def\iccvPaperID{5575} 

\ificcvfinal\fi

\begin{document}

\title{Dual Path Transformer with Partition Attention}

\author{Zhengkai Jiang\textsuperscript{\rm 1}\hspace{0.5em}Liang Liu\textsuperscript{\rm 1}\\Jiangning Zhang\textsuperscript{\rm 1}\hspace{0.5em}Yabiao Wang\textsuperscript{\rm 1}\hspace{0.5em}Mingang Chen\textsuperscript{\rm 2}\hspace{0.5em} Chengjie Wang\textsuperscript{\rm 1}\hspace{0.5em} \\
\textsuperscript{\rm 1}Tencent Youtu Lab, \textsuperscript{\rm 2}Shanghai Development Center of Computer Software Technolog\\
{\tt\small \{zhengkjiang, caseywang, jasoncjwang\}@tencent.com}\\}

\maketitle
\ificcvfinal\thispagestyle{empty}\fi

\begin{abstract}

This paper introduces a novel attention mechanism, called dual attention, which is both efficient and effective. The dual attention mechanism consists of two parallel components: local attention generated by Convolutional Neural Networks (CNNs) and long-range attention generated by Vision Transformers (ViTs). To address the high computational complexity and memory footprint of vanilla Multi-Head Self-Attention (MHSA), we introduce a novel Multi-Head Partition-wise Attention (MHPA) mechanism. The partition-wise attention approach models both intra-partition and inter-partition attention simultaneously. Building on the dual attention block and partition-wise attention mechanism, we present a hierarchical vision backbone called DualFormer. We evaluate the effectiveness of our model on several computer vision tasks, including image classification on ImageNet, object detection on COCO, and semantic segmentation on Cityscapes. Specifically, the proposed DualFormer-XS achieves 81.5\% top-1 accuracy on ImageNet, outperforming the recent state-of-the-art MPViT-XS by 0.6\% top-1 accuracy with much higher throughput.

\vspace{-0.6cm}
\end{abstract}

\section{Introduction}
In recent years, attention mechanisms~\cite{vaswani2017attention} and Transformers~\cite{krizhevsky2012imagenet} have emerged as powerful tools for architecture design. The use of Transformers, such as Vision Transformers (ViTs) originally introduced in natural language processing (NLP)~\cite{dosovitskiy2020image}, has become prevalent in architecture design for a variety of computer vision tasks due to their powerful long-range dependency modeling ability. Recent research~\cite{liu2021swin,dong2022cswin,lee2022mpvit,xu2021co} on Transformers has demonstrated promising performance on various computer vision tasks, including image classification~\cite{krizhevsky2012imagenet}, object detection~\cite{ren2015faster,dai2016r,lin2017feature,jiang2020learning,jiang2019video,jiang2022prototypical,jiang2023stc}, and semantic segmentation~\cite{long2015fully,chen2017deeplab}.


\begin{figure}[t]
\begin{center}
\includegraphics[width=0.98\linewidth]{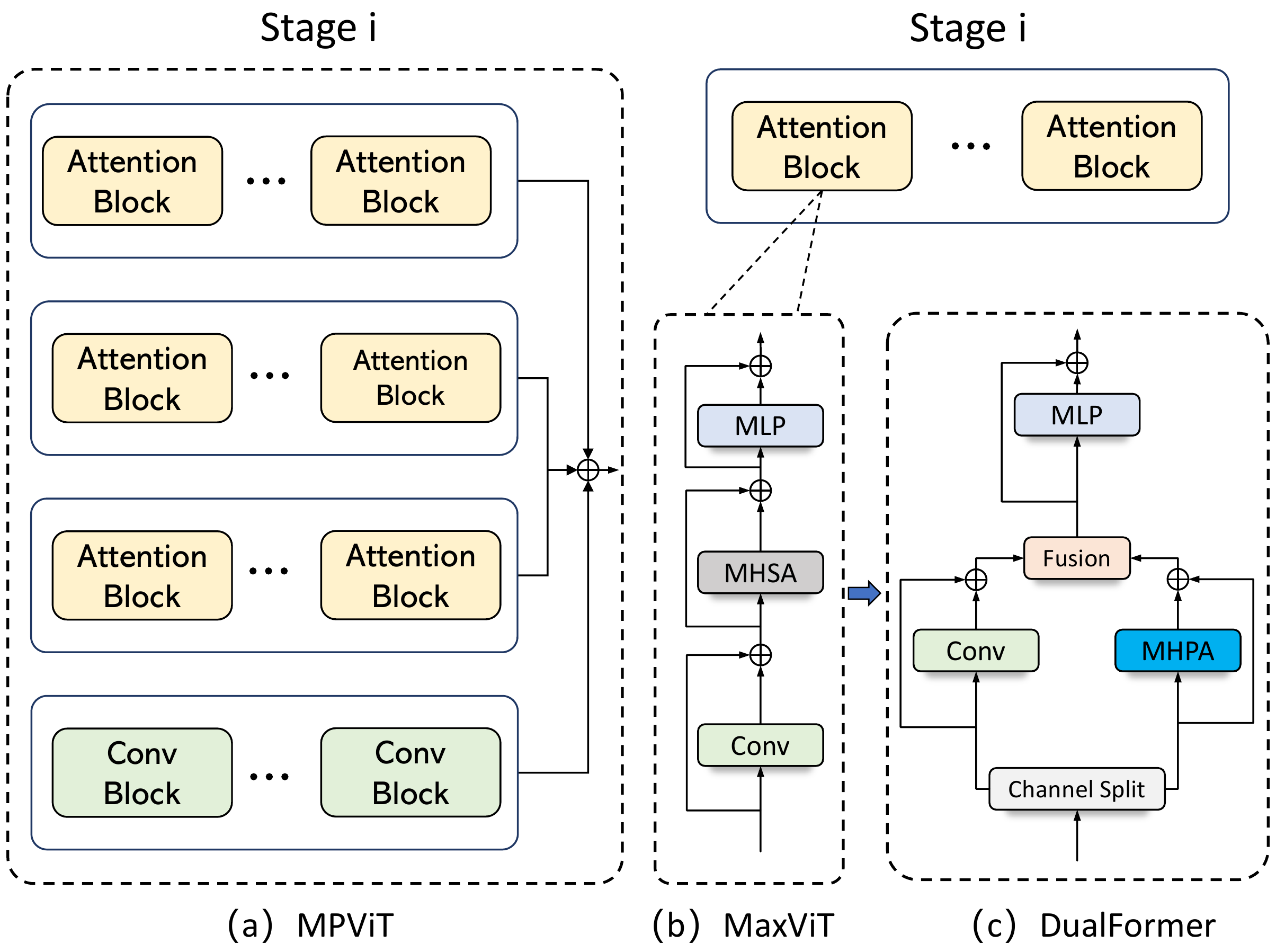}
\end{center}
\vspace{-0.2cm}
\caption{Comparison with recent ViT architectures, such as MPViT~\cite{lee2022mpvit}, which employs a stage-wise multi-path architecture in parallel, MaxViT~\cite{tu2022maxvit}, which stacks MBConv~\cite{sandler2018mobilenetv2} and self-attention blocks in series, and our proposed DualFormer, which efficiently combines partition-wise attention and MBConv blocks through a dual-path design.}
\label{fig:teaser}
\vspace{-0.4cm}
\end{figure}
Despite the remarkable success of visual transformers on various benchmarks, there still exist several challenges. The self-attention mechanism used in transformers results in quadratic computational and memory complexity, which limits their generalization on high-resolution images, particularly for dense prediction tasks such as object detection and semantic segmentation. To address this issue, several works propose self-attention variants, such as window attention~\cite{liu2021swin}, which is less effective to capture long-range dependencies, token selection~\cite{dong2022cswin, xia2022vision}, which relies on the heuristic rules and replacing early-stage self-attention with depth-wise convolution blocks~\cite{gao2021container, li2022uniformer} endowing local structure perception, which lacks long-range dependencies ability for low-level stages. Building upon previous work that has identified the existence of redundancy dependencies, as demonstrated in~\cite{cao2019gcnet}, and qualitative visualization seen in ~\ref{fig:partition_teaser}, we propose a novel technique called partition-wise attention that aims to achieve a balance between model performance and efficiency. The proposed method generates spatial partitions by clustering the feature representation level's similarity. Within each partition, the self-attention mechanism is performed, resulting in higher efficiency and a lower memory footprint. To enable the model to capture long-range dependencies, we also introduce inter-partition attention. 


Besides the efficiency issue, the ability to model multi-scale or multiple receptive fields is another crucial factor for downstream tasks. To endow visual transformers with multi-scale modeling ability, CoaT~\cite{xu2021co} proposes the co-scale mechanism, which represents fine and coarse features simultaneously. Similarly, MPViT~\cite{lee2022mpvit} introduces stage-wise multi-transformer paths in parallel to exploit multi-scale feature representation. However, both methods require heavy computation and memory overhead. In this paper, we propose \textit{DualFormer}, a simple and efficient \textit{dual path attention} mechanism to capture different scales and receptive field information through a novel approach. Our method first splits features along the channel dimension and then integrates the depth-wise convolution block and the proposed Multi-Head Partition-wise Attention (MHPA) in parallel before feeding them into the Feed-Forward Network (FFN) block. As illustrated in Fig~\ref{fig:framework}, DualFormer comprises several patch embeddings and dual attention blocks. Each dual attention block consists of a convolution branch that captures local-wise feature dependencies and a self-attention branch that captures global feature dependencies in parallel. The features from both branches are then aggregated to enable both fine and coarse feature representations.

To demonstrate the effectiveness and efficiency of our proposed method, we conducted experiments on various tasks, including image classification on ImageNet-1K dataset, object detection on the MSCOCO dataset, and semantic segmentation on ADE20K. Specifically, our DualFormer-S model, which has 22.6M parameters and 4.4G FLOPS, achieved 83.5\% top-1 accuracy for ImageNet-1K classification, 48.6\% mAP for MSCOCO object detection, and 48.6\% mIoU for ADE20K semantic segmentation. Our experiments demonstrate that DualFormer significantly outperforms state-of-the-art methods across different visual tasks

The main contributions are summarized as follows:  
\begin{itemize}

\item Dual Path Vision Transformer (\textit{DualFormer}) is proposed to simultaneously model various scales and receptive fields information, resulting in a more discriminative and detailed representation.

\item To address the complexity and memory issues associated with the standard self-attention block, we propose performing self-attention within each group partition and cross-attention in a group-wise mechanism to model long-range dependency.

\item Our proposed DualFormer achieves new state-of-the-art performances on a variety of vision tasks, including image classification, object detection, and semantic segmentation. 

\end{itemize}

\begin{figure}[t]
\begin{center}
\includegraphics[width=0.95\linewidth]{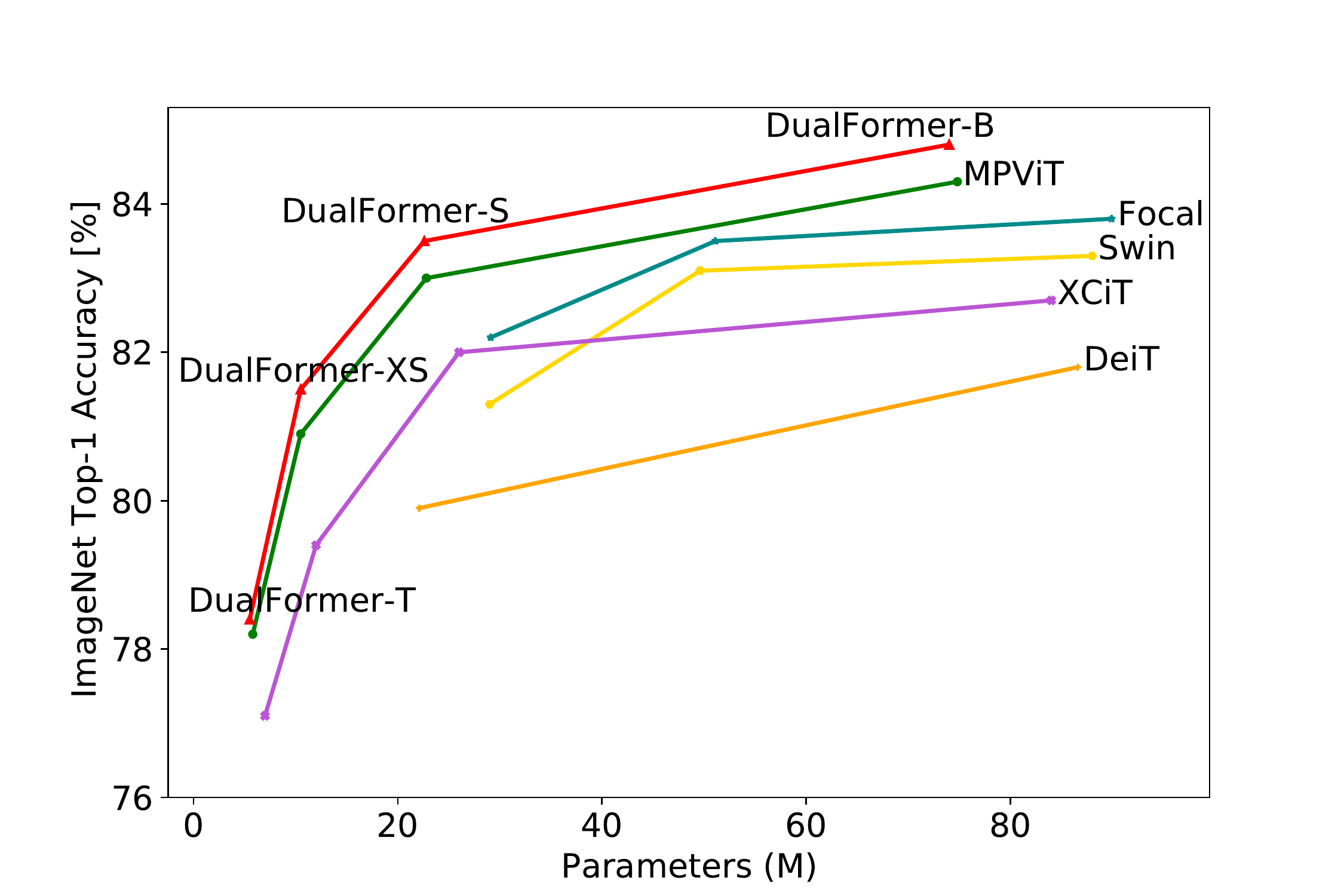}
\end{center}
\vspace{-0.1in}
\caption{\textbf{Parameters vs. ImageNet Accuracy}. DualFormers outperform state-of-the-art Vision Transformers while having fewer parameters and FLOPs. The model names, T, XS, S, and B, denote tiny, extra-small, small, and base, respectively.}
\label{fig:plot}
\vspace{-0.5cm}
\end{figure}

\begin{textblock}{43}(137,56.5)
  \begin{table*}[t]
  \centering
  \begingroup\setlength{\fboxsep}{0pt}
  \colorbox{white}{
  \begin{adjustbox}{width=\textwidth, center}
    \begin{tabular}{lccc}
    \textbf{Model} & \textbf{Top-1 Acc.} & \textbf{Param.} & \textbf{GFLOPs}\\
    \midrule
    ResNet-18~\cite{he2016deep} & 69.8 & ~~11.7M & ~~1.8 \\
    PVT-T~\cite{wang2021pyramid} & 75.1 & ~~13.2M & ~~1.9 \\
    CoaT Mi~\cite{xu2021co} & 78.9 & ~~10.0M & ~~6.8\\
    XCiT-T24/16~\cite{ali2021xcit} & 79.4 & ~~12.0M & ~~2.3\\
    MPViT-XS~\cite{lee2022mpvit} & 80.9 & ~~10.5M & ~~2.9 \\
    \midrule
    \textbf{DualFormer-XS~(ours)} & \textbf{81.5} & ~~10.5M & ~~2.3\\
    \end{tabular}
    \end{adjustbox}
    }
    \endgroup
  \label{tab:test}%
\end{table*}%
    \end{textblock}
\section{Related Work}

\subsection{Transformers for Vision}
Transformers have recently achieved remarkable success in visual recognition~\cite{dosovitskiy2020image,liu2021swin}, becoming the de-facto standard in natural language processing tasks. ViT~\cite{dosovitskiy2020image}, as the pioneering work that introduced Transformers into vision tasks, applies a standard Transformer to images by splitting an image into patches and providing the sequence of linear embeddings of those patches as input to a Transformer, achieving state-of-the-art performance on image classification tasks. DeiT~\cite{touvron2021training} introduces a distillation token mechanism that achieves competitive performance when trained only on ImageNet-1K with no external data. In addition to pure Transformer architectures, many researchers have explored combining CNNs and Transformers~\cite{gao2021container,xiao2021early,dai2021coatnet,li2022uniformer,park2022vision}. Container~\cite{gao2021container} unifies CNN and Transformer in a spatial context aggregation manner and further proposes to replace early-stage Multi-Head Self-Attention (MHSAs) with convolutions, exploiting the inductive bias of local convolutions in shallow layers and leading to faster convergence speeds. Similarly, Uniformer~\cite{li2022uniformer} adopts a similar approach to Container, stacking convolutions in shallow layers and self-attention in deep layers, addressing both redundancy and dependency for efficient and effective representation learning. Many works~\cite{xiao2021early,park2022vision} follow the paradigm of inserting convolutions in the early stage, which increases optimization stability and convergence speed, together with better performance. However, the quadratic complexity of Transformers still remains intractable for high-resolution images, particularly for dense prediction tasks such as object detection and semantic segmentation.

\subsection{Efficient Self-Attentions}
Since the image resolution of vision tasks is typically very high, developing an efficient self-attention scheme is critical. Recent works~\cite{liu2021swin,dong2022cswin} adopt the local self-attention mechanism and achieve global interaction through shifted windows scheme to reduce the quadratic complexity of Transformers. Efficient Attention~\cite{shen2021efficient} proposes a novel efficient self-attention mechanism by switching the order of query, key, and value, resulting in substantially lower memory and computational costs. Performer~\cite{choromanski2020rethinking} proposes a novel fast attention approach based on positive orthogonal random features with linear complexity. SOFT~\cite{lu2021soft} replaces the original softmax operation in self-attention with a Gaussian kernel function, yielding dot-product similarity through low-rank matrix decomposition. XCiT~\cite{ali2021xcit} proposes modeling interactions across feature channels rather than tokens, resulting in linear complexity in the number of tokens. HamNet~\cite{geng2021attention} formulates modeling global context as a low-rank recovery problem through matrix decomposition, outperforming a variety of attention modules on both image classification and dense prediction tasks with linear complexity. Flash Attention~\cite{dao2022flashattention} proposes an IO-aware exact attention algorithm using tiling to reduce the number of memory reads and writes, resulting in a dramatic speedup on the long-range sequence and better performance. In this paper, inspired by GCNet~\cite{cao2019gcnet} which shows that global contexts modeled by non-local networks are nearly the same for different query positions, we propose partition attention with a semantically dynamic group mechanism, discarding full self-attention, leading to high efficiency, especially for large resolution input images in object detection and semantic segmentation.

\begin{figure*}[!ht]
\centering
\includegraphics[width=0.92\textwidth]{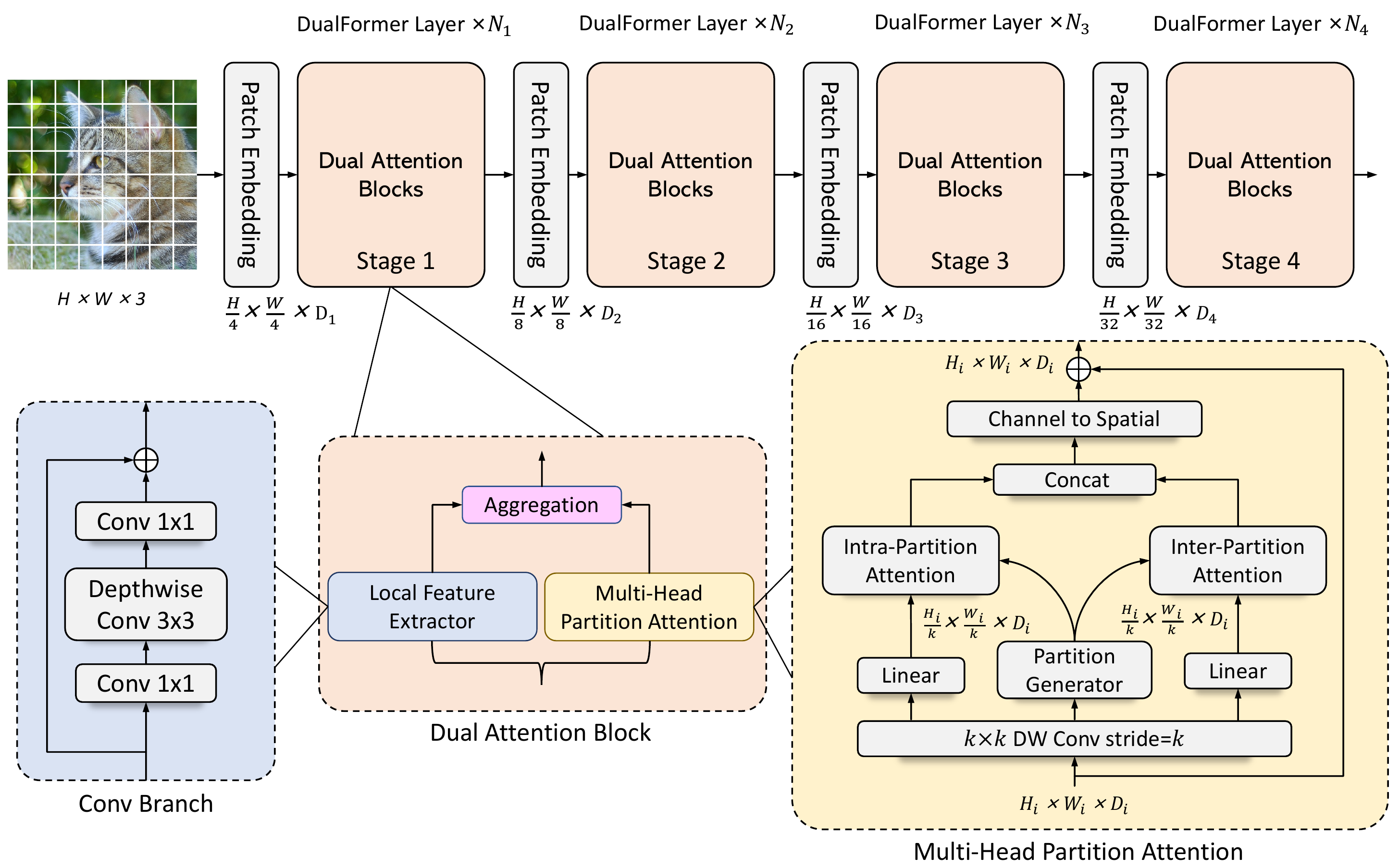}
\caption{\textbf{DualFormer architecture.} 
We adopt a typical hierarchical design, similar to previous works~\cite{lee2022mpvit,liu2021swin,wang2021pvtv2}, based on a proposed basic building block called the Dual Attention Block. This block unifies partition-wise attention and local feature extraction and generates four stages of features for downstream dense prediction tasks. To extract local features, we employ the MBConv block~\cite{sandler2018mobilenetv2}, which uses depth-wise convolution to capture local spatial interactions. The Multi-head Partition Attention (MHPA) generates different spatial partitions first and then performs intra-partition and inter-partition attention, which are fused to generate fine-grained global interaction features. To reduce the computational complexity, MHPA utilizes depth-wise stride convolution to downsample the input resolution and recovers the original resolution via a channel-to-spatial mechanism.}
\label{fig:framework}
\vspace{-0.5cm}
\end{figure*}

\subsection{Combination of CNN and ViT}
Another topic related to our work is the integration of Vision Transformers (ViT) and Convolutional Neural Networks (CNN) for general architecture design. Previous studies~\cite{gao2021container,li2022uniformer,pan2022integration} propose replacing self-attention in shadow layers with convolution, rather than combining them. CeiT~\cite{yuan2021incorporating} suggested using image-to-tokens to generate low-level features with convolutions, which enhances the locality and long-range dependency modeling of Transformers. CMT~\cite{guo2022cmt,dong2022cswin} introduced depth-wise convolution into the feed-forward network to encode local features. CvT~\cite{wu2021cvt} utilized convolutional token embedding before each stage and convolution projection of Transformer blocks. Subsequent works~\cite{xiao2021early,dai2021coatnet} combined convolution into early stage of transformers for improved generalization and scalability. MPViT~\cite{lee2022mpvit} proposes to use stage-wise multi-scale patch embeddings to exploit multi-scale and multi-path representation but accompanying high computational complexity. Inception Transformer~\cite{si2022inception} presents an inception mixer with high- and low-frequency paths based on a hand-crafted channel-splitting manner. The most related work is ACMix~\cite{pan2022integration} combining traditional convolution and self-attention in a hybrid way. However, both the self-attention branch and the convolution branch rely on the projected queries, keys, and values features with $1\times1$ convolutions. Inspired by HRNet~\cite{wang2020deep} success, we instead design parallel convolution and attention branch mechanisms to exploit different frequencies and various scales representation, leading the more discriminative features with better performance and higher efficiency.

\section{Method}
Vanilla self-attention has quadratic computational and memory complexity with respect to input resolution, which hinders its use for dense prediction tasks such as object detection and semantic segmentation. To address the efficiency issue of self-attention, we propose an alternative partition attention with a dynamic token group mechanism. This mechanism can model global dependencies with high efficiency. Additionally, to reduce the computational complexity of shadow layers of partition attention, we adopt stride depth-wise convolution to downsample the input resolution. Recent studies~\cite{tu2022maxvit,zhang2023vitaev2, yang2022moat, yu2022metaformer, dong2022cswin}  on transformer architecture design have adopted the paradigm of serially stacking global self-attention and local convolution. Inspired by the success of the multi-path and multi-scale in HRNet~\cite{wang2020high} and Inception Network~\cite{si2022inception}, we propose a dual-path design that performs attention and local convolution in parallel, which allows for various scale modeling to obtain more discriminative feature representations.


In this section, we first present the architecture of our proposed DualFormer. Next, we provide a detailed explanation of partition-wise attention, following a brief review of the most typical multi-head self-attention, which comprises four components: the \textit{partition generator}, \textit{intra-partition attention}, \textit{inter-partition attention}, and \textit{local-global aggregation}. We then discuss various approaches to combining convolution and attention. Lastly, we provide a detailed description of the different configurations of the proposed DualFormer.

\subsection{Overall Architecture}
The overall architecture of the proposed DualFormer is depicted in Fig~\ref{fig:framework}. Given an input image with a resolution of $H \times W \times 3$, we follow previous works~\cite{gao2021container,li2022uniformer,liu2021swin} and employ two successive overlapped convolutional token embedding layers ($3 \times 3$ convolution layer with stride $2$) to obtain $\frac{H}{4} \times \frac{W}{4}$ patch tokens with a dimension of $D$. The entire network comprises four stages that generate a hierarchical representation for downstream dense prediction tasks. For each stage $i\in {1,2,3,4}$, DualFormer consists of $N_i$ sequential \textit{Dual Attention Blocks} while keeping the number of tokens constant. For the $i^{th}$ stage, the feature maps have $\frac{H}{2^i+1} \times \frac{W}{2^i+1}$ tokens, which is similar to both CNN backbones like ResNet~\cite{he2016deep} and prior Transformer backbones like Swin~\cite{liu2021swin}. For image classification tasks, we use global average pooling from the last stage and feed it to the classification head. For dense prediction tasks, such as object detection and semantic segmentation, all four stages of feature maps are fed into the task head.


\subsection{Approximating Attention with Clustering}

\noindent \textbf{Review Self-Attention.} Given a set of $n$ token sequence $\bm{X} \in \mathbb{R}^{n\times d}$ with $d$-dimensional vector for each token, self-attention~\cite{vaswani2017attention} aims to compute a weighted sum of the values based on the affinity of each token. Mathematically, it can be formulated as follows:
\begin{equation}
    \bm{Q} = \bm{X}\bm{W}_q, \bm{K}=\bm{X}\bm{W}_k, \bm{V}=\bm{X}\bm{W}_v,
\end{equation}
where $\bm{W}_q\in \mathbb{R}^{d\times d_e}$, $\bm{W}_k\in \mathbb{R}^{d\times d_e}$, and $\bm{W}_v \in \mathbb{R}^{d\times d_e}$ are learnable projection weights. The query-specific attention maps $\bm{A} = \frac{\bm{Q}\bm{K}^{T}}{\sqrt{d_e}} \in \mathbb{R}^{n \times n}$ can be obtained by the scaled dot-product of $\bm{Q}$ and $\bm{K}$. The whole global aggregation operation can be formulated as:
\begin{equation}
    \text{Attention}(\bm{Q}, \bm{K}, \bm{V}) = \texttt{softmax}(\frac{\bm{Q}\bm{K}^{T}}{\sqrt{d_e}}) \bm{V}.
\end{equation}

\begin{figure}[t]
\begin{center}
\includegraphics[width=0.9\linewidth]{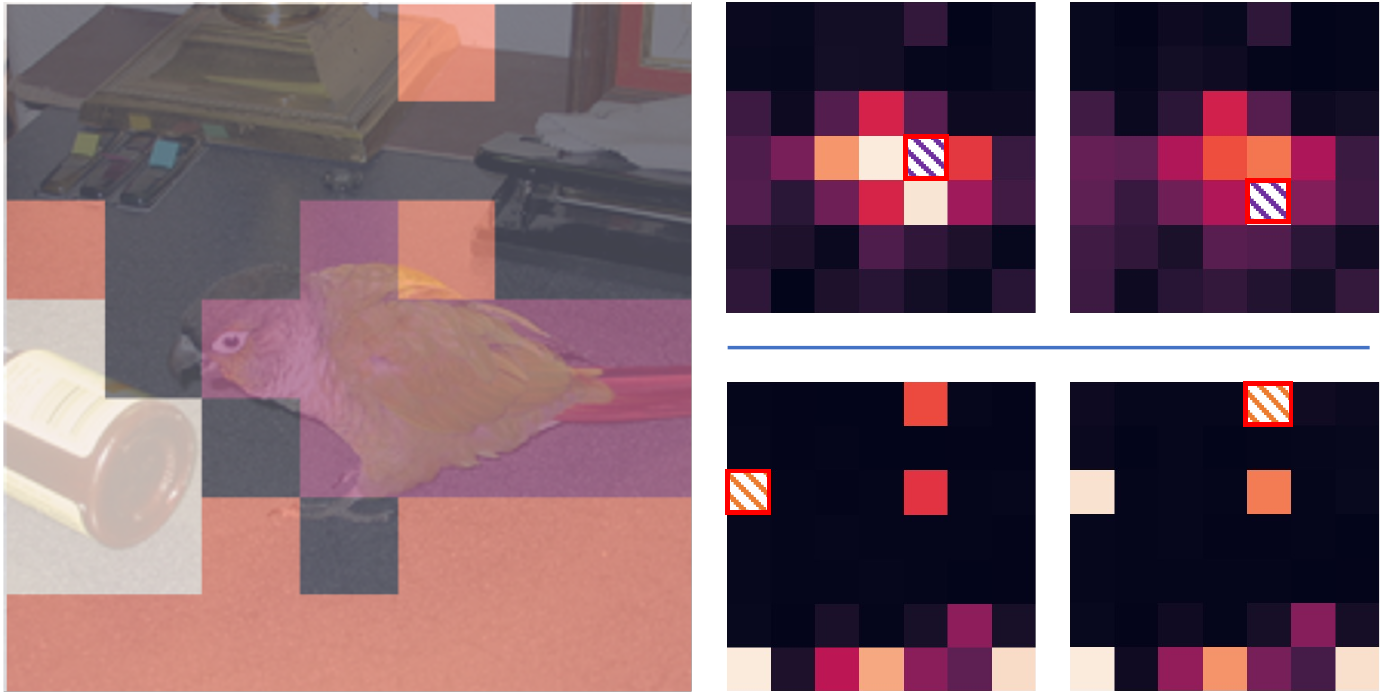}
\end{center}
\caption{Visualization of attention maps for various query positions (indicated by red boxes) in a self-attention block on ImageNet validation set. The original image (shown on the \textit{left}) is blended with a grouping result based on the feature map. It was observed that different queries within the same group exhibit similar attention maps.}
\label{fig:partition_teaser}
\vspace{-0.4cm}
\end{figure}
Through an analysis of the attention weights of a ViT model pre-trained on ImageNet, as shown in Fig~\ref{fig:partition_teaser}, we observe that the attention weights of different query locations are almost the same, indicating a large amount of redundancy. Thus phenomenon has also been observed in the GCNet~\cite{cao2019gcnet}. To address this issue, we propose a clustering-based partition attention mechanism, which is a fast approximation of self-attention. Clustering-based partition attention masks use of similarities between queries and groups them to reduce the computational cost. It mainly comprises four modules: the Partition Generator, Intra-Partition Attention, Inter-Partition Attention, and Aggregation.


\noindent \textbf{Partition Generator.} Given flatten input feature map $\bm{x} \in \mathbb{R}^{n \times d}$, we aim to group spatial locations into $K$ distinct clusters defining the partition $\mathcal{C} = \left \{ \bm{C}_k \right \}_{k=1}^{K}$, with $\cup_k \bm{C}_k=\{x_i \}_{i=1}^{n}$ and $\bm{C}_i\cap \bm{C}_j = \emptyset, \forall i\neq j$. Each cluster $\bm{C}_k$ of the partition is represented by a centroid $\mu_k \in \mathbb{R}^d, \forall k \in \{1,\dots, K\}$. The goal is to find centroids minimizing the following error function:
\begin{equation}
\begin{aligned}
& \min_{\mathcal{C}, \mu_{1},\ldots, \mu_K}
& & \sum_{k=1}^{K} \sum_{x_i \in C_k}  \mathcal{L}(x_i, \mu_k),
\end{aligned}
\end{equation}

To assign $x_i \in \mathbb{R}^d$, $i \in {1,\dots, n}$ to a cluster $\bm{C}_k$, we evaluate the similarity measure $\mathcal{L}$. Specifically, $x_i$  belongs to the cluster $\bm{C}_l$ if and only if $l = \argmin_k \mathcal{L}(x_i, \mu_k)$. The Euclidean distance in terms of their $\mathcal{L}_2$ norm can be used as the similarity measure, i.e., $\mathcal{L}(x_i, \mu_k)= \|x_i - \mu_k \|_2^2$. To improve the efficiency of solving the optimization problem, we use the more efficient and GPU-friendly Locality-Sensitive Hashing (LSH) algorithm ~\cite{dasgupta2011fast,kitaev2020reformer} as the default clustering algorithm. LSH leverages hash collisions to form different spatial groups. We define the hash function as follows:
\begin{equation}
    h_i(x_i)=1 \text{ if } \beta_i \cdot x_i \geq 0 \text{ else } 0,
\end{equation}
where $\cdot$ denotes the dot product and $x_i,\beta_i \in \mathbb{R}^{d}$ are the feature vector and the pre-defined norm vector, respectively. To divide all feature vectors into four groups for example, we randomly initialize two norm vectors $\beta_1, \beta_2$, and we assign each vector $x_i$ to a hash value:
\begin{equation}
 H(x)=h_1(x)+2h_2(x) \in \{0, 1, 2, 3\}.   
\end{equation}


This LSH scheme is efficient to implement on GPU in a batch of vectors mechanism. It is worth noting that we are not the first to use the LSH algorithm to improve the Transformer block. Reformer~\cite{kitaev2020reformer} replaces the dot-product attention with the fast LSH, reducing the complexity from $O(L^2)$ to $O(L\log L)$ where $L$ is the length of the sequence. Compared to our method, Reformer is applied to the NLP tasks while our work focuses on vision tasks. Furthermore, Reformer~\cite{kitaev2020reformer} still computes the dot-product attention inside the local groups, while DualFormer completely discards the dot-product attention mechanism, resulting in faster training and inference speed with less loss of accuracy.

\noindent \textbf{Intra-Partition Attention.} Let $\mathcal{I} = \left \{ I_k \right \}_{k=1}^{K}$ denote the clustered partition coordinates generated by the above partition generator. The intra-partition attention weight on the location $i$ which belongs to the cluster $I_k$ is defined as follows:
\begin{equation}
    w_i = \frac{x_i}{\sum_{j \in I_k} x_j}
\end{equation}
where $x_i$ represents the input feature at location $i$. To obtain the transformed input feature map $\widetilde x \in \mathbb{R}^{n \times d}$, we use an additional lightweight linear layer. The intra-partition attention feature $x^{intra}$ after communication within each partition is then calculated as:

\begin{equation}
   x_i^{intra} = \frac{w_i \cdot \widetilde x_i}{\sum_{j \in I_k} w_j}
\end{equation}

\noindent \textbf{Inter-Partition Attention.} In addition to capturing intra-partition interactions, inter-partition attention enables the modeling of long-range dependencies by achieving spatial interaction across each partition. To obtain the global descriptor $x^{global} \in \mathbb{R}^{K \times d}$ of all partitions, we use an average pooling operation by computing the mean of all the tokens in each cluster $I_k$:

\begin{equation}
   x_k^{global} = \frac{1}{l_k} \sum_{i \in I_k} \widetilde x_i
\end{equation}
Here, $l_k$ represents the number of tokens in cluster $I_k$. To achieve partition-level interactions, we predict the important coefficients $m \in \mathbb{R}^{K}$ for each partition, similar to the approach used in SENet~\cite{hu2018squeeze}. Finally, we compute the inter-partition interaction feature $x^{inter} \in \mathbb{R}^{K\times d}$ as follows:
\begin{equation}
   x_k^{inter} = x_k^{global} \cdot \frac{m_k}{\sum_{j \in I_k} m_j} 
\end{equation}
Here, $x_k^{inter}$ represents the interaction feature of the $k^{th}$ partition, $x_k^{global}$ is the global descriptor of $k^{th}$ partition,, and $m_k$ is the predicted importance coefficient for the $k^{th}$ partition. The normalization term $\sum_{j \in I_k} m_j$ ensures that the sum of all importance coefficients across all partitions is equal to 1.

\noindent \textbf{Global-to-Local Aggregation.}
To address the shape misalignment between intra-attention and inter-attention features, we first scatter the inter-partition interaction feature $x^{inter} \in \mathbb{R}^{K\times d}$ to 
$x^{intra} \in \mathbb{R}^{n\times d}$ based on the clustered groups $\mathcal{C} = \left \{ C_k \right \}_{k=1}^{K}$. After this, we aggregate the inter-attention feature $x^{intra}$ and $x^{inter}$ through concatenation to enhance global-wise dependencies, followed by a simple convolution layer.

\noindent \textbf{Channel to Spatial.} 
To reduce the computational cost, we downsample the input resolution as early stage from $H_i \times W_i \times D_i$ to $\frac{H_i}{k} \times \frac{W_i}{k} \times D_i$, where $k$ is the downsample rate. To recover the original spatial resolution, we use a $1\times1$ convolution to increase channel dimension from $D_i$ to $D_i \times k \times k$. The resulting tensor is then reshaped from $\frac{H_i}{k} \times \frac{W_i}{k} \times (D_i \times k \times k)$ to $H_i \times W_i \times D_i$. Additionally, we use a skip-connection with the original input features to preserve important details and avoid significant loss of information.

\subsection{Dual-Attention Transformer}
While many studies have focused on combining convolution and attention, few have investigated whether to stack convolution and attention blocks in a serial or parallel way. MaxViT~\cite{tu2022maxvit} proposes to stack MBConv, block, and grid attention serially, while Uniformer~\cite{li2022uniformer} replaces the attention in the shadow layers with convolution. Meanwhile, the multi-path structure allows for capturing different scales and receptive field attention, which has been successful in downstream tasks~\cite{wang2020high}. Therefore, we revisit the design mechanism of combining convolution and attention within each block and propose to stack them in parallel.

We implement our proposed dual attention block as a basic build block, similar to MSAs in the vanilla Transformer. As shown in Fig~\ref{fig:framework}, our dual-attention transformer block consists of a convolution block and a multi-head partition attention block. We adopt MBConv as the default convolution. The convolution block is composed of a $1\time 1$ convolution and depth-wise convolution, which aims to capture local structure information. The multi-head partition attention block aims to capture global-wise information. For a fair comparison with other ViTs, we build four different network configurations for our Dual-Transformers: DualFormer-T (Tiny), DualFormer-XS (Extra Small), DualFormer-S (Small), and DualFormer-B (Base).

\begin{table}[t]
  \centering
 \begin{adjustbox}{width=\columnwidth, center}
\scalebox{0.85}{
\begin{tabular}{l|cccccc}
\toprule
MPViT & \#Layers & Channels & Param. & GFLOPs \\
\midrule
Tiny~(T)     & [2, 2, 4, 2] & [~~64, ~~128, ~~256, ~~320]  & ~~5.5M   & ~~1.3 \\
XSmall~(XS)    & [2, 2, 4, 2] & [~~64,~~128, ~~320, ~~368]   & ~~10.5M  & ~~2.3  \\
Small~(S)     & [4, 4, 7, 3] & [~~64, ~~128, ~~320, ~~512]  & ~~22.6M  & ~~4.4  \\
Base~(B)    & [6, 12, 25, 7] & [~~64, ~~128, ~~368, ~~560]   & ~~74.0M  & ~~15.8  \\
\bottomrule
\end{tabular}%
}
\end{adjustbox}
  \caption{\textbf{DualFormer Configurations.} \#Layers and Channels denote the number of transformer encoders and the embedding dimension in each stage, respectively. FLOPs are measured using $224\times224$ input image.}
  \label{tab:arch}%
  \vspace{-0.5cm}
\end{table}%

\section{Experiments}
In this section, we evaluate the proposed DualFormer on three different tasks: image classification on ImageNet~\cite{deng2009imagenet}, semantic segmentation on ADE20K~\cite{zhou2019semantic}, and object detection and instance segmentation on COCO~\cite{lin2014microsoft}. Additionally, we conduct ablation studies to demonstrate the effectiveness of each component design.

\begin{table}[t]
  \centering
\begin{adjustbox}{width=\columnwidth, center}
  \scalebox{0.80}{
    \begin{tabular}{lcclc}
    \toprule
    Model & Param.(M) & GFLOPs & Top-1 & Reference\\
    \midrule
    DeiT-T~\cite{touvron2021training} & ~~5.7   & ~~1.3   & ~~72.2 & ICML21\\
    XCiT-T12/16~\cite{ali2021xcit} & ~~7.0   & ~~1.2   & ~~77.1 & NeurIPS21 \\
    CoaT-Lite T~\cite{xu2021co} & ~~5.7   & ~~1.6   & ~~76.6 & ICCV21 \\
    MPViT-T~\cite{lee2022mpvit} & ~~5.8   & ~~1.6   & ~~78.2 & CVPR22 \\
    \rowcolor{mycyan} \textbf{DualFormer-T} & ~~5.5   & ~~1.3   & \textbf{~~78.4} & \\
    \midrule
    ResNet-18~\cite{he2016deep} & ~~11.7  & ~~1.8   & ~~69.8 &CVPR16 \\
    PVT-T~\cite{wang2021pyramid} & ~~13.2  & ~~1.9   & ~~75.1 & ICCV21 \\
    XCiT-T24/16~\cite{ali2021xcit} & ~~12.0  & ~~2.3   & ~~79.4 & NeurIPS21 \\
    CoaT Mi~\cite{xu2021co} & ~~10.0  & ~~6.8   & ~~80.8 & ICCV21 \\
    MPViT-XS~\cite{lee2022mpvit} & ~~10.5  & ~~2.9   & ~~80.9 & CVPR22 \\
    PVT-ACmix-T~\cite{pan2022integration} & ~~13.2 & ~~2.0  & ~~78.0 & CVPR22\\ 
    \rowcolor{mycyan} \textbf{DualFormer-XS} & ~~10.5  & ~~2.3   & \textbf{~~81.5}& \\
    \midrule
    ResNet-50~\cite{he2016deep} & ~~25.6  & ~~4.1   & ~~76.1 &CVPR16 \\
    PVT-S~\cite{wang2021pyramid} & ~~24.5  & ~~3.8   & ~~79.8 & ICCV21 \\
    DeiT-S/16~\cite{touvron2021training} & ~~22.1  & ~~4.6   & ~~79.9 & ICML21 \\
    Swin-T~\cite{liu2021swin} & ~~29.0  & ~~4.5   & ~~81.3 & ICCV21 \\
    CvT-13~\cite{wu2021cvt} & ~~20.0  & ~~4.5   & ~~81.6 & ICCV21 \\
    XCiT-S12/16~\cite{ali2021xcit} & ~~26.0  & ~~4.8   & ~~82.0 & NeurIPS21 \\
    Focal-T~\cite{yang2021focal} & ~~29.1  & ~~4.9   & ~~82.2 & NeurIPS21 \\
    CoaT S~\cite{xu2021co} & ~~22.0  & ~~12.6  & ~~82.1 & ICCV21 \\
    CrossViT-18~\cite{chen2021crossvit} & ~~43.3 & ~~9.5 & ~~82.8 & ICCV21 \\
    CoaT-Lite S~\cite{xu2021co} & ~~20.0  & ~~4.0   & ~~81.9 & ICCV21 \\
    MPViT-S~\cite{lee2022mpvit} & ~~22.8  & ~~4.7   & ~~83.0 &  CVPR22 \\
    iFormer-S~\cite{si2022inception} & ~~20.0 &~~4.8 & ~~83.4 & NeurIPS22 \\
    \rowcolor{mycyan} \textbf{DualFormer-S} & ~~22.6  & ~~4.4   & \textbf{~~83.5} & \\
    \midrule
    ResNeXt-101~\cite{xie2017aggregated} & ~~83.5  & ~~15.6  & ~~79.6 &CVPR17 \\
    PVT-L~\cite{wang2021pyramid} & ~~61.4  & ~~9.8   & ~~81.7 & ICCV21 \\
    MaxViT-S\cite{tu2022maxvit} & ~~69.0 & ~~11.7 & ~~ 84.5 & ECCV22 \\
    DeiT-B/16~\cite{touvron2021training} & ~~86.6  & ~~17.6  & ~~81.8 & ICML21 \\
    XCiT-M24/16~\cite{ali2021xcit} & ~~84.0  & ~~16.2  & ~~82.7 & NeurIPS21 \\
    Swin-B~\cite{liu2021swin} & ~~88.0  & ~~15.4  & ~~83.3 & ICCV21 \\
    XCiT-S12/8~\cite{ali2021xcit} & ~~26.0  & ~~18.9  & ~~83.4 & NeurIPS21 \\
    Focal-B~\cite{yang2021focal} & ~~89.8  & ~~16.0  & ~~83.8  & NeurIPS21 \\
    MPViT-B~\cite{lee2022mpvit} & ~~74.8  & ~~16.4  & ~~84.3 & CVPR22  \\
    iFormer-B~\cite{si2022inception} & ~~87.0 & ~~14.0 & ~~84.6 & NeurIPS22\\
    \rowcolor{mycyan} \textbf{DualFormer-B} & ~~74.0  & ~~15.8  & \textbf{~~84.8}&  \\
    \bottomrule
    \end{tabular}
    }
    \end{adjustbox}
 \caption{\textbf{ImageNet-1K classification.} These models are trained with $224\times224$ resolution. For fair comparison, we do not include models that are distilled~\cite{touvron2021training} or use $384\times384$ resolution.}
  \label{tab:cls}
  \vspace{-0.5cm}
\end{table}

\begin{table*}[htbp]
  \centering
      \begin{adjustbox}{max width=\textwidth}\
\begin{tabular}{lll|cccccc|cccccc}
    \toprule
    \multirow{2}[2]{*}{Backbone} & \multirow{2}[2]{*}{Params. (M)} & \multirow{2}[2]{*}{GFLOPs}  & \multicolumn{6}{c|}{Mask R-CNN $1\times$ schedule} & \multicolumn{6}{c}{Mask R-CNN $3\times$ schedule + MS} \\
          &       &             & $AP^b$ & $AP^b_{50}$ & $AP^b_{75}$ & $AP^m$ & $AP^m_{50}$ & $AP^m_{75}$ & $AP^b$ & $AP^b_{50}$ & $AP^b_{75}$ & $AP^m$ & $AP^m_{50}$ & $AP^m_{75}$ \\
    \midrule
    ResNet18~\cite{he2016deep} & ~~31 & ~~207 & 34.0 & 54.0 & 36.7 & 31.2 & 51.0 & 32.7 & 36.9 & 57.1 & 40.0 & 33.6 & 53.9 & 35.7 \\
    PVT-T~\cite{wang2021pyramid} & ~~33 & ~~240 & 36.7 & 59.2 & 39.3 & 35.1 & 56.7 & 37.3 & 39.8 & 62.2 & 43.0 & 37.4 & 59.3 & 39.9 \\
    MPViT-T~\cite{lee2022mpvit} & ~~28 & ~~216 & 42.2 & 64.2 & 45.8 & 39.0 & 61.4 & 41.8 & 44.8 & 66.9 & 49.2 & 41.0 & 64.2 & 44.1  \\
    \rowcolor{mycyan} \textbf{DualFormer-T} & ~~25 & ~~208 & 42.4 & 65.0 & 46.4 & 39.2 & 61.6 & 42.2 & 45.1 & 67.3 & 49.6 & 41.2 & 64.3 & 44.3 \\
    \midrule
    ResNet50~\cite{he2016deep} & ~~44 & ~~260 & 38.0 & 58.6 & 41.4 & 34.4 & 55.1 & 36.7 & 41.0 & 61.7 & 44.9 & 37.1 & 58.4 & 40.1 \\
    PVT-S~\cite{wang2021pyramid} & ~~44 & ~~245 & 40.4 & 62.9 & 43.8 & 37.8 & 60.1 & 40.3 & 43.0 & 65.3 & 46.9 & 39.9 & 62.5 & 42.8 \\
    Swin-T~\cite{liu2021swin} & ~~48 & ~~264 & 43.7 & 66.6 & 47.7 & 39.8 & 63.3 & 42.7 & 46.0 & 68.1 & 50.3 & 41.6 & 65.1 & 44.9 \\
    Focal-T~\cite{yang2021focal} & ~~49 & ~~291 & 44.8 & 67.7 & 49.2 & 41.0 & 64.7 & 44.2 & 47.2 & 69.4 & 51.9 & 42.7 & 66.5 & 45.9 \\
    MPViT-XS~\cite{lee2022mpvit} & ~~30 & ~~231 & 44.2 & 66.7 & 48.4 & 40.4 & 63.4 & 43.4 & 46.6 & 68.5 & 51.1 & 42.3 & 65.8 & 45.8  \\
    
    \rowcolor{mycyan} \textbf{DualFormer-XS} & ~~29 & ~~219 & 44.6 & 67.1 & 48.9 & 40.8 & 63.9 & 43.8 & 47.0 & 69.0 & 51.6 & 42.5 & 66.1 & 45.8   \\
    \midrule
    ResNet101~\cite{he2016deep} & ~~63 & ~~336 & 40.4 & 61.1 & 44.2 & 36.4 & 57.7 & 38.8 & 42.8 & 63.2 & 47.1 & 38.5 & 60.1 & 41.3 \\
    PVT-M~\cite{wang2021pyramid} & ~~64 & ~~392 & 42.0 & 64.4 & 45.6 & 39.0 & 61.6 & 42.1 & 44.2 & 66.0 & 48.2 & 40.5 & 63.1 & 43.5 \\
    Swin-S~\cite{liu2021swin} & ~~69 & ~~359 & 46.5 & 68.7 & 51.3 & 42.1 & 65.8 & 45.2 & 48.5 & 70.2 & 53.5 & 43.3 & 67.3 & 46.6 \\
    Focal-S~\cite{yang2021focal} & ~~71 & ~~401 & 47.4 & 69.8 & 51.9 & 42.8 & 66.6 & 46.1 & 48.8 & 70.5 & 53.6 & 43.8 & 67.7 & 47.2  \\
    MPViT-S~\cite{lee2022mpvit} & ~~43 & ~~268 & 46.4 & 68.6 & 51.2 & 42.4 & 65.6 & 45.7 & 48.4 & 70.5 & 52.6 & 43.9 & 67.6 & 47.5  \\
   
    \rowcolor{mycyan} \textbf{DualFormer-S} & ~~43 & ~~258 & 46.8 & 69.0 & 51.5 & 42.6 & 66.0 & 45.9 & 48.6 & 70.5 & 52.8 & 44.0 & 67.7 & 47.3 \\
    \midrule
    ResNeXt101-64x4d~\cite{xie2017aggregated} & ~~102 & ~~493 & 42.8 & 63.8 & 47.3 & 38.4 & 60.6 & 41.3 & 44.4 & 64.9 & 48.8 & 39.7 & 61.9 & 42.6 \\
    PVT-L~\cite{wang2021pyramid} & ~~81 & ~~457 & 42.9 & 65.0 & 46.6 & 39.5 & 61.9 & 42.5 & 44.5 & 66.0 & 48.3 & 40.7 & 63.4 & 43.7 \\
    Swin-B~\cite{liu2021swin} & ~~107 & ~~496 & 46.9 & 69.2 & 51.6 & 42.3 & 66.0 & 45.5 & 48.5 & 69.8 & 53.2 & 43.4 & 66.8 & 49.6 \\
    Focal-B~\cite{yang2021focal} & ~~110 & ~~533 & 47.8 & 70.2 & 52.5 & 43.2 & 67.3 & 46.5 & 49.0 & 70.1 & 53.6 & 43.7 & 67.6 & 47.0   \\
    MPViT-B~\cite{lee2022mpvit} & ~~95 & ~~503 & 48.2 & 70.0 & 52.9 & 43.5 & 67.1 & 46.8 & 49.4 & 70.9 & 54.3 & 44.5 & 68.1 & 48.2  \\
    
    \rowcolor{mycyan} \textbf{DualFormer-B} & ~~95 & ~~495 & 48.5 & 70.3 & 53.0 & 43.6 & 67.2 & 46.9 & 49.6 & 71.0 & 54.5 & 44.6 & 68.2 & 48.4 \\
    \bottomrule
    \end{tabular}
    \end{adjustbox}
    \caption{\textbf{COCO detection and instance segmentation} with Mask R-CNN~\cite{he2017mask}. Models are trained for$1\times$ schedule and $3\times$ schedule~\cite{wu2019detectron2} with multi-scale training inputs~(MS)~\cite{liu2021swin,sun2021sparse}. 
    All backbones are pretrained on ImageNet-1K. For fair comparison, we omit models pretrained on larger-datasets~(\eg, ImageNet-21K). The GFLOPs are measured at resolution $800 \times 1280$.}
  \label{tab:det}
  \vspace{-0.5cm}
\end{table*}

\subsection{Image Classification}
\noindent\textbf{Dataset.} We evaluate our proposed DualFormer on the ImageNet-1K dataset~\cite{deng2009imagenet}, which contains 1.2 million training images and 50,000 validation images with 1,000 semantic categories. We adopt the AdamW~\cite{loshchilov2017decoupled} optimizer with an initial learning rate of $5\times 10^{-4}$, momentum of 0.9, and weight decay of $5\times 10^{-2}$. The batch size is 1024, and the default number of epochs is 300, trained on 8 Tesla V100 GPUs. During training, we set the number of linear warm-up and cool-down epochs to 5 and 10, respectively. In other epochs, we decrease the learning rate with a cosine annealing schedule. Following previous work~\cite{touvron2021training}, we use data augmentation techniques such as random flipping, mixup~\cite{zhang2018mixup}, and cutmix~\cite{yun2019cutmix}. We report the Top-1 accuracy under the single crop setting, which is a common evaluation metric. Additionally, we report the model size and the number of floating-point operations to display the trade-off between accuracy and model size.

\noindent\textbf{Results.} 
Table~\ref{tab:cls} presents the Top-1 accuracy achieved by DualFormer in the image classification task, in comparison to previous state-of-the-art models. Notably, DualFormer outperforms most ConvNets, ViTs, and MLPs with similar parameters and computational costs. For instance, DualFormer-T achieves higher accuracy than the recent MPViT-XS~\cite{lee2022mpvit} by 0.2\%, while using 18.8\% fewer FLOPs. Moreover, DualFormer-S demonstrates a significant improvement in Top-1 accuracy, with gains of 1.1\% and 2.2\% compared to Focal Transformer~\cite{yang2021focal}. 

\subsection{Object Detection}
\noindent\textbf{Dataset.} We conducted an evaluation of DualFormer on the COCO 2017 benchmark~\cite{lin2014microsoft} for both object detection and instance segmentation tasks. The COCO 2017 dataset comprises 118K train images and 5k validation images. To ensure a fair comparison, we followed the train and validation recipe of PVT~\cite{wang2021pyramid} for both object detection and instance segmentation. The backbone was pre-trained on ImageNet-1k, and the training was performed using a batch size of 16 on 8 Tesla V100 GPUs. The number of training epochs was set to 12 and 36 (1$\times$ schedule), following the methodology of previous works~\cite{he2017mask,liu2021swin,lee2022mpvit}. 

\noindent\textbf{Results.} Table~\ref{tab:det} presents the results of our evaluation of COCO 2017 using Mask R-CNN. Our DualFormer model outperforms pure ConvNets such as ResNet~\cite{he2016deep}, as well as Transformer variants PVT~\cite{wang2021pyramid} and Swin Transformer~\cite{liu2021swin} across all metrics. Notably, DualFormer-S consistently outperforms Swin-T~\cite{liu2021swin} by approximately 0.9\% in box AP and 1.0\% in mask AP of Mask R-CNN under 1$\times$ evaluation, while using significantly fewer parameters and FLOPs.
\begin{table}[t]
  \centering
  \begin{adjustbox}{width=0.80\columnwidth, center}
  \scalebox{0.65}{
    \begin{tabular}{lccc}
    \toprule
    Backbone & Params. & GFLOPs    & mIoU \\
    \midrule
    Swin-T~\cite{liu2021swin} & ~~59M   & ~~945     & 44.5 \\
    Focal-T~\cite{yang2021focal} & ~~62M   & ~~998      & 45.8 \\
    XCiT-S12/16~\cite{ali2021xcit} & ~~54M   & ~~966     & 45.9 \\
    XCiT-S12/8~\cite{ali2021xcit} & ~~53M   & 1237    & 46.6 \\
    MPViT-S~\cite{lee2022mpvit} & ~~52M & ~~943  & 48.3 \\
    iFormer-S~\cite{si2022inception} & ~~49M & ~~938  & 48.4 \\
    \rowcolor{mycyan} \textbf{DualFormer-S} & ~~48M & ~~922 & 48.6 \\
    \midrule
    ResNet-101~\cite{he2016deep} & 85M  &1029 &43.8         \\
    XCiT-S24/16~\cite{ali2021xcit} & ~~76M   & 1053    & 46.9 \\
    Swin-S~\cite{liu2021swin} & ~~81M   & 1038    & 47.6 \\
    XCiT-M24/16~\cite{ali2021xcit} & 112M  & 1213    & 47.6 \\
    Focal-S~\cite{yang2021focal} & ~~85M   & 1130     & 48.0 \\
    Swin-B~\cite{liu2021swin} & 121M  & 1841    & 48.1 \\
    XCiT-S24/8~\cite{ali2021xcit} & ~~74M   & 1587     & 48.1 \\
    XCiT-M24/8~\cite{ali2021xcit} & 110M  & 2161     & 48.4 \\
    Focal-B~\cite{yang2021focal} & 126M  & 1354     & 49.0 \\
    MPViT-B~\cite{lee2022mpvit} &   105M   &   1186 & 50.3 \\
    \rowcolor{mycyan} \textbf{DualFormer-B} & 104M & 1150 & 50.5\\
    \bottomrule
    \end{tabular}
    }
\end{adjustbox}
  \caption{\textbf{ADE20k semantic segmentation} results using UperNet~\cite{xiao2018unified}. GFLOPs are calculated with resolution $512\times2048$. For a fair comparison, We do not include models that are pre-trained on larger datasets~(\textit{i.e.,} ImageNet-21K).}
  \label{tab:seg}
  \vspace{-0.6cm}
\end{table}%

\subsection{Semantic Segmentation}
\noindent\textbf{Dataset.} The ADE20K~\cite{zhou2019semantic} dataset is a widely used benchmark for semantic segmentation, comprising 150 object and stuff classes and diverse scene annotations. The dataset includes 20K images in the training set and 2k images in the validation set. To conduct our experiments, we utilized the MMSegmentation~\cite{chen2019mmdetection} toolbox~\cite{mmseg2020} as our codebase. To ensure a fair comparison, we followed the training and validation recipe of PVT~\cite{wang2021pyramid} for semantic segmentation experiments. We employed UperNet~\cite{xiao2018unified} as the segmentation head, and the backbones were initialized with weights pre-trained on ImageNet-1K. To optimize our model, we utilized an AdamW~\cite{loshchilov2017decoupled} optimizer with an initial learning rate of $10^{-4}$ and weight decay of $10^{-4}$. The training was performed using a batch size of 32 on 8 Tesla V100 GPUs, and a total of 80K iterations were trained.

\noindent\textbf{Results.} Table~\ref{tab:seg} displays the results of our evaluation on the UperNet model for semantic segmentation. Our DualFormer model achieves stable performance gains over MPViT~\cite{lee2022mpvit}, which incorporates a multi-path branch for architecture design. Notably, with approximately 50M parameters, DualFormer-S outperforms MPViT-S by 0.3\% in mIoU. Furthermore, compared to the recent work iFormer~\cite{si2022inception}, DualFormer-S surpasses iFormer-S by 0.2\% in mIoU while using fewer parameters and FLOPs, demonstrating the effectiveness of DualFormer's design.

\subsection{Ablation Study}

\noindent\textbf{Effectiveness of each component.} To demonstrate the effectiveness of each component, we conducted ablation studies on the ImageNet-1k dataset, and the results of DualFormer-XS are presented in Table~\ref{tab:component}. In this table, \textit{Parallel} refers to the stacking way of the convolution and self-attention, \textit{Intra} refers to the proposed intra-partition attention module, \textit{Inter} refers to the proposed inter-partition attention module, and \textit{Depth-wise} means whether using the depth-wise convolution of local feature extractor. We achieved an accuracy of 80.6\% with only the intra-partition attention. This accuracy further improved to 81.2\% when combining the intra- and inter-partition attention, demonstrating the effectiveness of long-range information interaction. When equipped with the depth-wise convolution branch, i.e., MBConv, DualFormer-XS achieved an accuracy of 81.5\%, obtaining a 0.3\% accuracy gain, which indicates the effectiveness of the convolution branch.

\begin{table}[t]
  \centering
  \begin{adjustbox}{width=0.95\columnwidth, center}
  \scalebox{0.95}{
\begin{tabular}{ccccc}
\toprule
Parallel & Depth-wise & Intra & Inter& Top-1 Acc \\
\midrule
\cmark & \xmark & \cmark& \xmark &  80.6 \\
\cmark & \xmark & \xmark & \cmark & 80.8 \\
\cmark & \xmark & \cmark & \cmark & 81.2 \\
\xmark & \cmark & \cmark & \cmark & 81.0 \\
\cmark & \cmark & \cmark & \cmark & 81.5 \\
\bottomrule
\end{tabular}}
\end{adjustbox}
  \caption{Ablation study on each component of DualFormer block. We report Top-1 accuracy based on DualFormer-XS. Intra and Inter mean intra-partition and inter-partition attention, respectively.}
\label{tab:component}
  \vspace{-0.4cm}
\end{table}

\noindent\textbf{Different Clustering Methods.} We conduct an ablation experiment to study the influence of different clustering methods, which is shown in Table~\ref{tab:lshvskmeans}. Both methods promote the baseline significantly. Compared with LSH~\cite{dasgupta2011fast}, K-Means~\cite{lloyd1982least} further increases the accuracy by a slight margin of 0.1\%, but with extreme speed drops. Therefore, we adopt LSH to strike an ideal balance between speed and accuracy.

\begin{table}[!ht]\centering
\resizebox{0.45\textwidth}{!}{
\begin{tabular}{ccccc}
\toprule
Method & Throughput & Param & GFLOPs & Top-1  \\
\midrule
MPViT-XS & 640 & 10.9 & 2.5 & 80.9 \\
K-Means & 806 & 10.5 & 2.3 & 81.6 \\
LSH & 1253 & 10.5 & 2.3 & 81.5 \\

\bottomrule
\end{tabular}}
\caption{Comparison of different clustering methods with MPViT-XS. The results are based on DualFormer-XS. We report throughput and Top-1 accuracy on the ImageNet-1K validation set.}
\label{tab:lshvskmeans}
\vspace{-0.3cm}
\end{table}

\begin{figure}[!ht]
\centering
\includegraphics[width=0.48\textwidth]{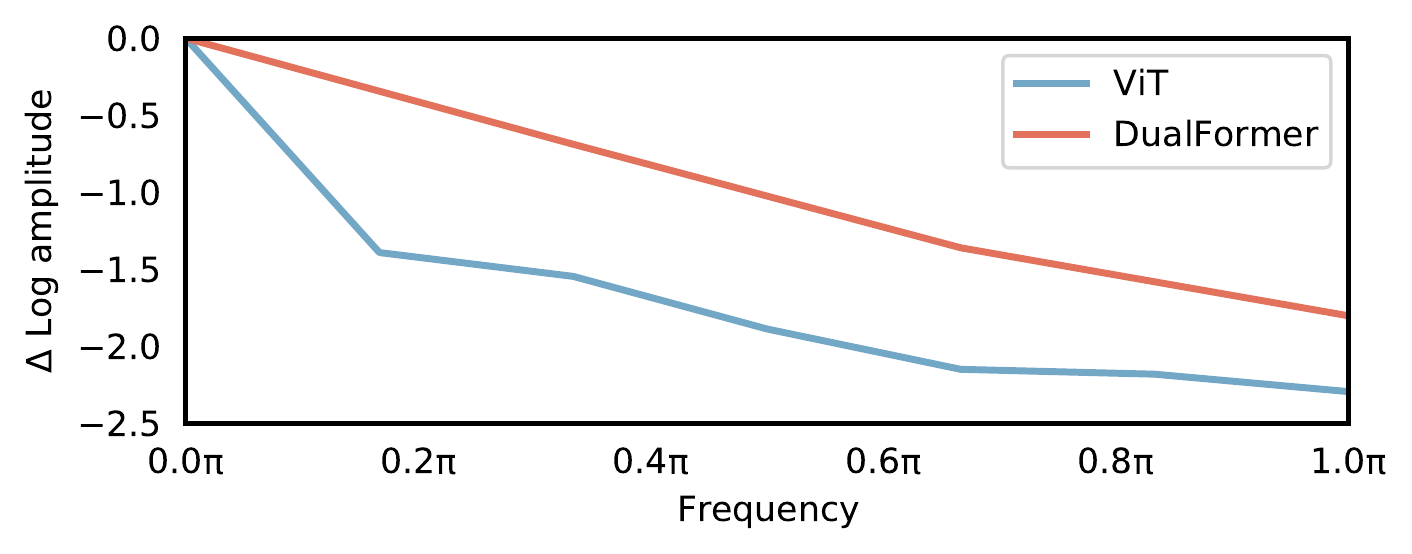}
\caption{Comparison of relative log amplitudes of Fourier
transformed feature maps for ViT and DualFormer.}
\label{fig:fourier}
\vspace{-0.5cm}
\end{figure}

\noindent\textbf{Fourier Analysis.} To exploit why dual-attention block works, we conduct visualization from a Fourier analysis following previous work~\cite{park2022vision}, which is shown in Fig~\ref{fig:fourier}. The results indicate that DualFormer has a greater amplitude at higher frequencies compared to vanilla ViT. This finding supports the hypothesis that DualFormer can capture more high-frequency information, which in turn helps to generate more robust and discriminative feature representations.

\noindent\textbf{Stacking convolution and self-attention in parallel or in series?} To explore the effective way of combining self-attention and convolution, we also implemented a stacking in-series approach, which is shown in Table~\ref{tab:component}. However, changing from the parallel approach to the series approach resulted in a 0.5\% Top-1 accuracy drop, indicating the effectiveness of the parallel approach. Additionally, compared with the recent stage-wise parallel approach MPViT as shown in Table~\ref{tab:lshvskmeans}, DualFormer excels both in accuracy and efficiency.

\begin{figure}[!ht]
\centering
\includegraphics[width=0.44\textwidth]{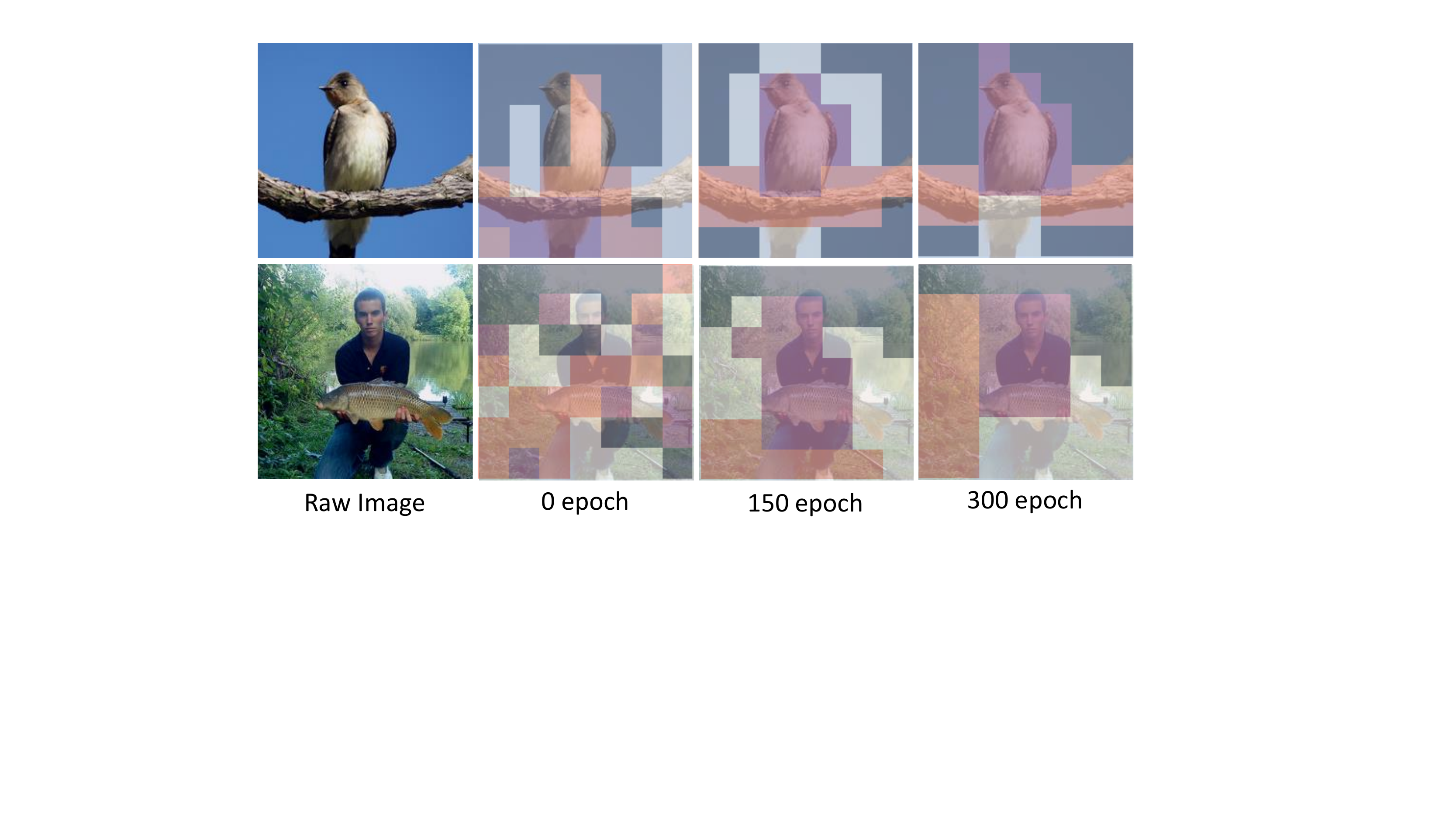}
\caption{Visualization of the partition result at different training epochs. Raw images are taken from ImageNet-1K validation set.}
\label{fig:partition}
\vspace{-0.3cm}
\end{figure}

\noindent\textbf{Partition Visualization.} To qualitatively analyze whether the clustering operation forms semantically meaningful groups, we visualized the partition results of three training periods, i.e., 0, 150, and 300 epochs. As shown in Fig~\ref{fig:partition}, the grouping process becomes more rational as the training epochs increase. The semantically similar positions are grouped together. For instance, in the first row of Fig~\ref{fig:partition}, the grouping process initially appears nonsensical. However, as the training progresses, the body of the bird is well-clustered into the purple group at 150 epochs, despite the white group wrongly classifying the tail of the bird as tokens of the sky. Ultimately, at 300 epochs, the positions of the tail are successfully separated into a new group. These results demonstrate that DualFormer is capable of capturing reasonable feature representations at different positions.

\section{Conclusion}
In this paper, we propose to combine convolution and attention in parallel to adaptively capture different scales and receptive field information, which is seldom adopted by previous works. To address the issue of spatial redundancy in vanilla self-attention, we introduce a clustering scheme to approximate the self-attention mechanism. The scheme involves partitioning the feature maps using local sensitivity hashing and computing the proxy features of each group. We then activate these proxy features through a global-wise interaction mechanism and distribute them to the corresponding query positions. Based on the proposed dual-attention block, we present DualFormer, which achieves state-of-the-art performance on various tasks including image classification, object detection, and semantic segmentation. We hope that our approach will inspire further research on effective and efficient ways of combining convolution and attention.

{\small
\bibliographystyle{ieee_fullname}
\bibliography{egbib}
}

\end{document}